\documentclass[runningheads]{llncs}
\usepackage{graphicx}
\usepackage{graphicx}
\usepackage{amsmath,amssymb}
\usepackage{makecell}
\usepackage{subfigure}
\usepackage[misc]{ifsym}
\usepackage{xcolor}
\usepackage{multirow}
\usepackage{hyperref}
\usepackage{booktabs}
\usepackage{algorithm}
\usepackage{algorithmic}
\usepackage{makecell}

\begin{document}
\title{DiffMIC: Dual-Guidance Diffusion Network\\ 
for Medical Image Classification}
\author{
Yijun Yang\inst{1,2}
\and
Huazhu Fu\inst{3}
\and
Angelica I Aviles-Rivero\inst{4}
\and
Carola-Bibiane Schönlieb\inst{4}
\and
Lei Zhu\inst{1,2,} $^\textrm{\Letter}$
}
%
\authorrunning{Y. Author et al.}
%
\institute{The Hong Kong University of Science and Technology (Guangzhou), China 
\email{leizhu@ust.hk}\\
\and
The Hong Kong University of Science and Technology, Hong Kong, China \and 
Institute of High Performance Computing, Agency for Science,
Technology and Research, Singapore
\and
University of Cambridge, UK\\
}
\maketitle              
\begin{abstract}
Diffusion Probabilistic Models have recently shown remarkable performance in generative image modeling, attracting significant attention in the computer vision community. However, while a substantial amount of diffusion-based research has focused on generative tasks, few studies have applied diffusion models to general medical image classification.
In this paper, we propose the first diffusion-based model (named DiffMIC) to address general medical image classification by eliminating unexpected noise and perturbations in medical images and robustly capturing semantic representation. To achieve this goal, we devise a dual conditional guidance strategy that conditions each diffusion step with multiple granularities to improve step-wise regional attention. Furthermore, we propose learning the mutual information in each granularity by enforcing Maximum-Mean Discrepancy regularization during the diffusion forward process.
We evaluate the effectiveness of our DiffMIC on three medical classification tasks with different image modalities, including placental maturity grading on ultrasound images, skin lesion classification using dermatoscopic images, and diabetic retinopathy grading using fundus images. Our experimental results demonstrate that DiffMIC outperforms state-of-the-art methods by a significant margin, indicating the universality and effectiveness of the proposed model.
Our code is publicly available at \href{https://github.com/scott-yjyang/DiffMIC}{https://github.com/scott-yjyang/DiffMIC}.

\end{abstract}
\section{Introduction}
\label{sec:intro}

Medical image analysis plays an indispensable role in clinical therapy because of the implications of digital medical imaging in modern healthcare~\cite{de2016machine}. Medical image classification, a fundamental step in the analysis of medical images, strives to distinguish medical images from different modalities based on certain criteria. An automatic and reliable classification system can help doctors interpret medical images quickly and accurately.
Massive solutions for medical image classification have been developed over the past decades in the literature, most of which adopt deep learning ranging from popular CNNs to vision transformers~\cite{Esteva2017,Esteva2019,Rajpurkar2022,Shamshad2022}.  These methods have the potential to reduce the time and effort required for manual classification and improve the accuracy and consistency of results. However, medical images with diverse modalities still challenge existing methods due to the presence of various ambiguous lesions and fine-grained tissues, such as ultrasound (US), dermatoscopic, and fundus images. Moreover, generating medical images under hardware limitations can cause noisy and blurry effects, which can degrade image quality and thus demand a more effective feature representation modeling for robust classifications.

Recently, Denoising Diffusion Probabilistic Models (DDPM)~\cite{ho2020denoising} have achieved excellent results in image generation and synthesis tasks~\cite{pmlr-v139-nichol21a,batzolis2021conditional,dhariwal2021diffusion,singh2022high} by iteratively improving the quality of a given image.
Specifically, DDPM is a generative model based on a Markov chain, which models the data distribution by simulating a diffusion process that evolves the input data towards a target distribution.
Although a few pioneer works tried to adopt the diffusion model for image segmentation and object detection tasks~\cite{amit2021segdiff,wolleb2022diffusion,chen2022diffusiondet,han2022card}, their potential for high-level vision has yet to be fully explored.

Motivated by the achievements of diffusion probabilistic models in generative image modeling, \textbf{1) we present a novel Denoising Diffusion-based model named DiffMIC} for accurate classification of diverse medical image modalities. As far as we know, we are the first to propose a Diffusion-based model for general medical image classification. Our method can appropriately eliminate undesirable noise in medical images as the diffusion process is stochastic in nature for each sampling step.
\textbf{2) In particular, we introduce a Dual-granularity Conditional Guidance (DCG) strategy} to guide the denoising procedure, conditioning each step with both global and local priors in the diffusion process. By conducting the diffusion process on smaller patches, our method can distinguish critical tissues with fine-grained capability. 
\textbf{3) Moreover, we introduce Condition-specific Maximum-Mean Discrepancy (MMD) regularization} to learn the mutual information in the latent space for each granularity, enabling the network to model a robust feature representation shared by the whole image and patches.
\textbf{4) We evaluate the effectiveness of DiffMIC on three 2D medical image classification tasks} including placental maturity grading, skin lesion classification, and diabetic retinopathy grading. The experimental results demonstrate that our diffusion-based classification method consistently and significantly surpasses state-of-the-art methods for all three tasks. 
\section{Method}
\label{sec:method}

\begin{figure*}[!t]
\centering
\includegraphics[width=1.0\textwidth]{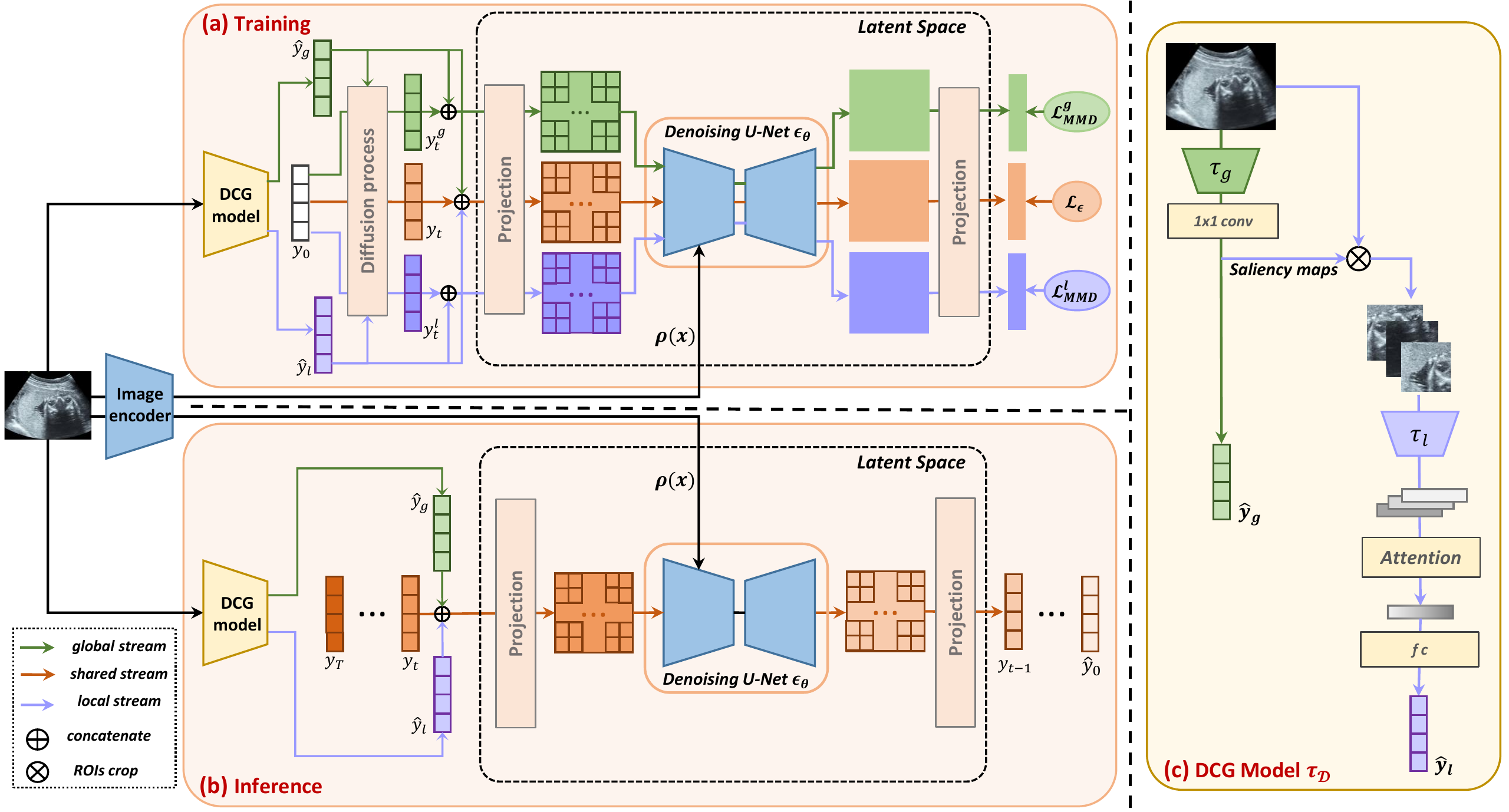}
\vskip -5pt
\caption{\textbf{Overview of our DiffMIC framework. } (a) The training phase (forward process) and (b) The inference phase (reverse process) are constructed, respectively. 
(The noise of feature embedding is greater with the darker color.) (c) The DCG Model $\tau_\mathcal{D}$ guides the diffusion process by the dual priors from the raw image and ROIs.}
\label{fig:overview} 
\end{figure*}

Figure~\ref{fig:overview} shows the schematic illustration of our network for medical image classification. 
Given an input medical image $x$, we pass it to an image encoder to obtain the image feature embedding $\rho(x)$, and a dual-granularity conditional guidance (DCG) model to produce the global prior $\hat{y}_{g}$ and local prior $\hat{y}_{l}$.
At the training stage, we apply the diffusion process on ground truth $y_0$ and different priors to generate three noisy variables $y^g_t$, $y^l_t$, and $y_t$ (the global prior for $y^g_t$, the local prior for $y^l_t$, and dual priors for $y_t$). 
Then, we combine the three noisy variables $y^g_t$, $y^l_t$, and $y_t$ and their respective priors and project them into a latent space, respectively. We further integrate three projected embeddings with the image feature embedding $\rho(x)$ in the denoising U-Net, respectively, and predict the noise distribution sampled for $y^g_t$, $y^l_t$, and $y_t$. 
We devise condition-specific maximum-mean discrepancy (MMD) regularization loss on the predicted noise of $y^g_t$ and $y^l_t$, and employ the noise estimation loss by mean squared error (MSE) on the predicted noise of $y_t$ to collaboratively train our DiffMIC network.

\vspace{3pt}
\noindent 
\textbf{Diffusion Model.} \ 
Following DDPM~\cite{ho2020denoising}, our diffusion model also has two stages: a forward diffusion stage (training) and a reverse diffusion stage (inference). 
In the forward process, the ground truth response variable $y_0$ is added Gaussian noise through the diffusion process conditioned by time step $t$ sampled from a uniform distribution of $[1, T]$, and such noisy variables are denoted as $\{y_1,...,y_t,..,y_T\}$.
As suggested by the standard implementation of DDPM, we adopt a UNet as the denoising network to parameterize the reverse diffusion process and learn the noise distribution in the forward process. 
In the reverse diffusion process, the trained UNet $\epsilon_\theta$ generates the final prediction $\hat{y}_0$ by transforming the noisy variable distribution $p_\theta(y_T)$ to the ground truth distribution $p_\theta(y_0)$: 
\begin{equation} 
p_\theta(y_{0:T-1}|y_T,\rho(x)) = \prod_{t=1}^{T}p_\theta(y_{t-1}|y_t,\rho(x)), \; \text{ and} \quad
p_\theta(y_T)=\mathcal{N}(\frac{\hat{y}_g+\hat{y}_l}{2},\mathbb{I}
),
\label{eq:mmd}
\end{equation} 
where $\theta$ is parameters of the denoising UNet, $\mathcal{N}(\cdot,\cdot)$ denotes the Gaussian distribution, and $\mathbb{I}$ is the identity matrix.

\subsection{Dual-granularity Conditional Guidance (DCG) Strategy}
\label{sec:dcg}
\textbf{DCG Model.}
In most conditional DDPM, the conditional prior will be a unique given information. 
However, medical image classification is particularly challenging due to the ambiguity of objects. It is difficult to differentiate lesions and tissues from the background, especially in low-contrast image modalities, such as ultrasound images. 
Moreover, unexpected noise or blurry effects may exist in regions of interest (ROIs), thereby hindering the understanding of high-level semantics. 
Taking only a raw image $x$ as the condition in each diffusion step will be insufficient to robustly learn the fine-grained information, resulting in classification performance degradation. 

To alleviate this issue, we design a Dual-granularity Conditional Guidance (DCG) for encoding each diffusion step. 
Specifically, we introduce a DCG model $\tau_\mathcal{D}$ to compute the global and local conditional priors for the diffusion process.
Similar to the diagnostic process of a radiologist, we can obtain a holistic understanding from the global prior and also concentrate on areas corresponding to lesions from the local prior when removing the negative noise effects.
As shown in Figure~\ref{fig:overview} (c), for the global stream, the raw image data $x$ is fed into the global encoder $\tau_g$ and then a $1\times1$ convolutional layer to generate a saliency map of the whole image. The global prior $\hat{y}_{g}$ is then predicted from the whole saliency map by averaging the responses. 
For the local stream, we further crop the ROIs whose responses are significant in the saliency map of the whole image. 
Each ROI is fed into the local encoder $\tau_l$ to obtain a feature vector. 
We then leverage the gated attention mechanism\cite{ilse2018attention} to fuse all feature vectors from ROIs to obtain a weighted vector, which is then utilized for computing the local prior $\hat{y}_{l}$ by one linear layer.

\vspace{2mm}
\noindent
\textbf{Denoising Model.}
The noisy variable $y_t$ is sampled in the diffusion process based on the global and local priors computed by the DCG model following:
\begin{equation}
    y_t = \sqrt{\bar{\alpha}_t}y_0+\sqrt{1-\bar{\alpha}_t}\epsilon+(1-\sqrt{\bar{\alpha}_t})(\hat{y}_{g}+\hat{y}_{l}),
\end{equation}
where $\epsilon \sim \mathcal{N}(0, I)$, $\bar{\alpha}_t=\prod_{t}\alpha_t, \alpha_t=1-\beta_t$ with a linear noise schedule $\{\beta_t\}_{t=1:T}\in (0,1)^T$. 
After that, we feed the concatenated vector of the noisy variable $y_t$ and dual priors into our denoising model UNet $\epsilon_\theta$ to estimate the noise distribution, which is formulated as:
\begin{equation}
    \epsilon_\theta(\rho(x), y_t, \hat{y}_{g}, \hat{y}_{l}, t) = D(E(f([y_t,\hat{y}_{g},\hat{y}_{l}]), \rho(x),t),t),
\end{equation}
where $f(\cdot)$ denotes the projection layer to the latent space. $[\cdot]$ is the concatenation operation. $E(\cdot)$ and $D(\cdot)$ are the encoder and decoder of UNet. 
Note that the image feature embedding $\rho(x)$ is further integrated with the projected noisy embedding in the UNet to make the model focus on high-level semantics and thus obtain more robust feature representations.
In the forward process, we seek to minimize the noise estimation loss $\mathcal{L}_\epsilon$:
\begin{equation}
    \mathcal{L}_\epsilon = ||\epsilon-\epsilon_\theta(\rho(x), y_t, \hat{y}_{g}, \hat{y}_{l}, t)||^2 .
\end{equation}
Our method improves the vanilla diffusion model by conditioning each step estimation function on priors that combine information derived from the raw image and ROIs. 

\subsection{Condition-specific MMD Regularization}
%
Maximum-Mean Discrepancy (MMD) is to measure the similarity between two distributions by comparing all of their moments~\cite{gretton2006kernel,li2015generative}, which can be efficiently achieved by a kernel function. 
Inspired by InfoVAE~\cite{zhao2017infovae}, we introduce an additional pair of condition-specific MMD regularization loss to learn mutual information between the sampled noise distribution and the Gaussian distribution. 
To be specific, we sample the noisy variable $y_t^g$ from the diffusion process at time step $t$ conditioned only by the global prior and then compute an MMD-regularization loss as:
\begin{equation}
    \begin{aligned}
         \mathcal{L}^{g}_{MMD}(n||m) & =  \mathbb{K}(n,n^{'})-2\mathbb{K}(m,n)+\mathbb{K}(m,m^{'}), \\
   \text{with}   \;\;  &n=\epsilon, \;\;  m = \epsilon_\theta(\rho(x),\sqrt{\bar{\alpha}_t}y_0+\sqrt{1-\bar{\alpha}_t}\epsilon+(1-\sqrt{\bar{\alpha}_t})\hat{y}_{g},\hat{y}_{g},t),  
    \end{aligned}
    \label{eq:mmdg}
\end{equation} 
where $\mathbb{K}(\cdot, \cdot)$ is a positive definite kernel to reproduce distributions in the Hilbert space. The condition-specific MMD regularization is also applied on the local prior, as shown in Figure~\ref{fig:overview} (a).
While the general noise estimation loss $\mathcal{L}_\epsilon$ captures the complementary information from both priors, the condition-specific MMD regularization maintains the mutual information between each prior and target distribution.
This also helps the network better model the robust feature representation shared by dual priors and converge faster in a stable way.

\subsection{Training and Inference Scheme}

\vspace{2mm}
\noindent
\textbf{Total loss.} \
By adding the noise estimation loss and the MMD-regularization loss, we compute the total loss $\mathcal{L}_{diff}$ of our denoising network as follows:
\begin{equation}
    \mathcal{L}_{diff} = \mathcal{L}_\epsilon + \lambda(\mathcal{L}^{g}_{MMD}+\mathcal{L}^{l}_{MMD}),
\end{equation}
where $\lambda$ is a balancing hyper-parameter, and it is empirically set as $\lambda$=0.5.

%
\vspace{2mm}
\noindent
\textbf{Training details.} \ The diffusion model in this study leverages a standard DDPM training process, where the diffusion time step $t$ is selected from a uniform distribution of $[1, T]$, and the noise is linearly scheduled with $\beta_1 = 1\times10^{-4}$ and $\beta_T = 0.02$. 
We adopt ResNet18 as the image encoder $\rho(\cdot)$. 
Following \cite{han2022card}, we concatenate $y_t$,$\hat{y}_{g}$,$\hat{y}_{l}$, and apply a linear layer with an output dimension of 6144 to obtain the fused vector in the latent space. 
To condition the response embedding on the timestep, we perform a Hadamard product between the fused vector and a timestep embedding. We then integrate the image feature embedding and response embedding by performing another Hadamard product between them. The output vector is sent through two consecutive fully-connected layers, each followed by a Hadamard product with a timestep embedding. Finally, we use a fully-connected layer to predict the noise with an output dimension of classes. It is worth noting that all fully-connected layers are accompanied by a batch normalization layer and a Softplus non-linearity, with the exception of the output layer.
For the DCG model $\tau_D$, the backbone of its global and local stream is ResNet. We adopt the standard cross-entropy loss as the objective of the DCG model. We jointly train the denoising diffusion model and DCG model after pretraining the DCG model 10 epochs for warm-up, thereby resulting in an end-to-end DiffMIC for medical image classification.

\vspace{2mm}
\noindent
\textbf{Inference stage.} \
As displayed in Figure~\ref{fig:overview} (b), given an input image $x$, we first feed it into the DCG model to obtain dual priors $\hat{y}_g, \hat{y}_l$. 
Then, following the pipeline of DDPM, the final prediction $\hat{y}_0$ is iteratively denoised from the random prediction $y_T$ using the trained UNet conditioned by dual priors $\hat{y}_g, \hat{y}_l$ and the image feature embedding $\rho(x)$.

\section{Experiments}\label{sec:ex}

\noindent \textbf{Datasets and Evaluation:}
We evaluate the effectiveness of our network on an in-home dataset and two public datasets, e.g., PMG2000, HAM10000~\cite{tschandl2018ham10000}, and APTOS2019~\cite{aptos2019-blindness-detection}.
\textbf{(a) PMG2000.}  We collect and annotate a benchmark dataset (denoted as PMG2000) for placental maturity grading (PMG) with four categories\footnote{Our data collection is approved by the Institutional Review Board (IRB).}. 
PMG2000 is composed of 2,098 ultrasound images, and we randomly divide the entire dataset into a training part and a testing part at an 8:2 ratio.
\textbf{(b) HAM10000.}  
HAM10000~\cite{tschandl2018ham10000} is from the Skin Lesion Analysis Toward Melanoma Detection 2018 challenge, and it contains 10,015 skin lesion images with predefined 7 categories. 
\textbf{(c) APTOS2019.}  In APTOS2019~\cite{aptos2019-blindness-detection}, 
A total of 3,662 fundus images have been labeled to classify diabetic retinopathy into five different categories.
Following the same protocol in~\cite{gong2020distractor},  we split HAM10000 and APTOS2019 into a train part and a test part at a 7:3 ratio. 
These three datasets are with different medical image modalities. PMG2000 is gray-scale and class-balanced ultrasound images; HAM10000 is colorful but class-imbalanced dermatoscopic images; and APTOS2019  is another class-imbalanced dataset with colorful Fundus images. 
Moreover, we introduce two widely-used metrics Accuracy and F1-score to quantitatively compare our framework against existing SOTA methods. 

\noindent \textbf{Implementation Details:}
Our framework is implemented with the PyTorch on one NVIDIA RTX 3090 GPU. 
We center-crop the image and then resize the spatial resolution of the cropped image to $224$$\times$$224$. Random flipping and rotation for data augmentation are implemented during the training processing.
In all experiments, we extract six $32$$\times$$32$ ROI patches from each image.
%
We trained our network end-to-end using the batch size of 32 and the Adam optimizer.
The initial learning rate for the denoising model U-Net is set as 1×10$^{-3}$, while for the DCG model (see Section~\ref{sec:dcg}) it is set to 2×10$^{-4}$ when training the entire network. 
Following~\cite{marrakchi2021fighting}, the number of training epochs is set as 1,000 for all three datasets. 
In inference, we empirically set the total diffusion time step $T$ as 100 for PMG2000, 250 for HAM10000, and 60 for APTOS2019, which is much smaller than most of the existing works~\cite{ho2020denoising,han2022card}.
The average running time of our DiffMIC is about 0.056 seconds for classifying an image with a spatial resolution of $224$$\times$$224$.

\begin{table*}[!tbp]
\caption{Quantitative comparison to SOTA methods on three classification tasks. The best results are marked in bold font.}

\centering
\subtable[PMG2000]{
\vspace{-3mm}
    \resizebox{\textwidth}{!}{%
\begin{tabular}{c|c|ccccc|c} 
\toprule
\multicolumn{2}{c|}{Methods}                  & ResNet~\cite{he2016deep} & ViT~\cite{dosovitskiy2020image}   & Swin~\cite{liu2021swin}  & PVT~\cite{wang2021pyramid}   & GMIC~\cite{shen2021interpretable}   & \textbf{Our DiffMIC}  \\ 
\hline
\multirow{2}{*}{PMG2000} & Accuracy & 0.879  & 0.886 & 0.893 & 0.907 & 0.900  & \textbf{0.931}       \\
                                   & F1-score & 0.881  & 0.890 & 0.892 & 0.902 & 0.901 & \textbf{0.926}       \\
\bottomrule

\end{tabular}
\label{tab:main1}
    }}

\subtable[HAM10000 and APTOS2019]{

    \resizebox{\textwidth}{!}{%
\begin{tabular}{c|c|cccccc|c} 
\toprule
\multicolumn{2}{c|}{Methods}          & LDAM~\cite{cao2019learning}  & OHEM~\cite{shrivastava2016training}  & MTL~\cite{liao2018deep}   & DANIL~\cite{gong2020distractor} & CL~\cite{marrakchi2021fighting}    & ProCo~\cite{yang2022proco} & \textbf{Our DiffMIC}  \\ 
\hline
\multirow{2}{*}{HAM10000}  & Accuracy & 0.857 & 0.818 & 0.811 & 0.825 & 0.865 & 0.887 & \textbf{0.906}          \\
                           & F1-score & 0.734 & 0.660 & 0.667 & 0.674 & 0.739 & 0.763 & \textbf{0.816}          \\ 
\hline
\multirow{2}{*}{APTOS2019} & Accuracy & 0.813 & 0.813 & 0.813 & 0.825 & 0.825 & 0.837 & \textbf{0.858}          \\
                           & F1-score & 0.620 & 0.631 & 0.632 & 0.660 & 0.652 & 0.674 & \textbf{0.716}          \\
\bottomrule
\end{tabular}
\label{tab:main2}
    }}

\label{tab:main}
\vskip -5pt
\end{table*}

\noindent \textbf{Comparison with State-of-the-art Methods:}   
In Table~\ref{tab:main1}, we compare our DiffMIC against many state-of-the-art CNNs and transformer-based networks, including ResNet, Vision Transformer (ViT), Swin Transformer (Swin), Pyramid Transformer (PVT), and a medical image classification method (i.e., GMIC) on PMG2000. 
Apparently, PVT has the largest Accuracy of 0.907, and the largest F1-score of 0.902 among these methods.
More importantly, our method further outperforms PVT. 
It improves the Accuracy from 0.907 to 0.931, and the F1-score from 0.902 to 0.926.

Note that both HAM10000 and APTOS2019 have a class imbalance issue. 
Hence, we compare our DiffMIC against state-of-the-art long-tailed medical image classification methods, and report the comparison results in Table~\ref{tab:main2}.
For HAM10000, our method produces a promising improvement over the second-best method ProCo of 0.019 and 0.053 in terms of Accuracy and F1-score, respectively. 
For APTOS2019, our method obtains a considerable improvement over ProCo of 0.021 and 0.042 in Accuracy and F1-score respectively.

\begin{table*}[!tbp]
\centering
  \caption{Effectiveness of each module in our DiffMIC on the PMG2000 dataset. 
  }
  \vskip -5pt
    \resizebox{0.65\textwidth}{!}{%
\begin{tabular}{c|ccc|cc} 
\toprule
 & Diffusion & DCG & MMD-reg & Accuracy & F1-score \\
\hline
basic    &     \textbf{-}               &       \textbf{-}       &       \textbf{-}           & 0.879             & 0.881             \\
C1          & \checkmark         &        \textbf{-}      &      \textbf{-}            & 0.906             & 0.899             \\
C2          & \checkmark         & \checkmark   &    \textbf{-}              & 0.920             & 0.914             \\
\hline
\textbf{Our method}        & \checkmark         & \checkmark   & \checkmark       & \textbf{0.931}    & \textbf{0.926}    \\
\bottomrule
\end{tabular}
    }
\label{tab:ablation}
\vskip -5pt
\end{table*}

\begin{figure*}[!t]
    \centering
    \resizebox{0.85\textwidth}{!}{%
    \subfigure{\rotatebox{90}{\scriptsize{~~~PMG2000}}
        \includegraphics[width=\textwidth]{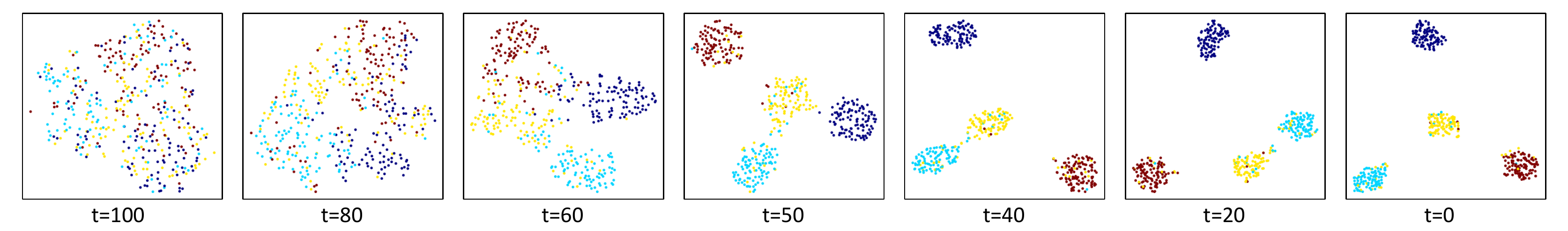}\label{fig:pmg}}\vspace{-10pt}}
    \resizebox{0.85\textwidth}{!}{%
    \subfigure{\rotatebox{90}{\scriptsize{~~~HAM10000}}
        \includegraphics[width=\textwidth]{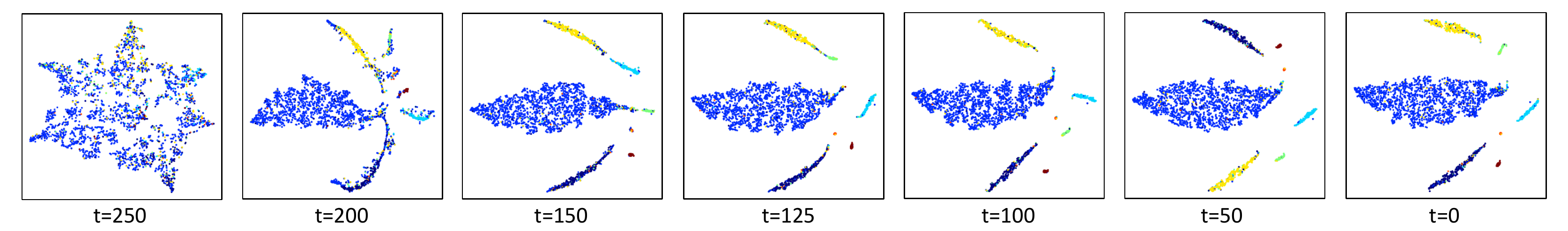}\label{fig: ham}}\vspace{-10pt}}
    \resizebox{0.85\textwidth}{!}{%
    \subfigure{\rotatebox{90}{\scriptsize{~~APTOS2019}}
        \includegraphics[width=\textwidth]{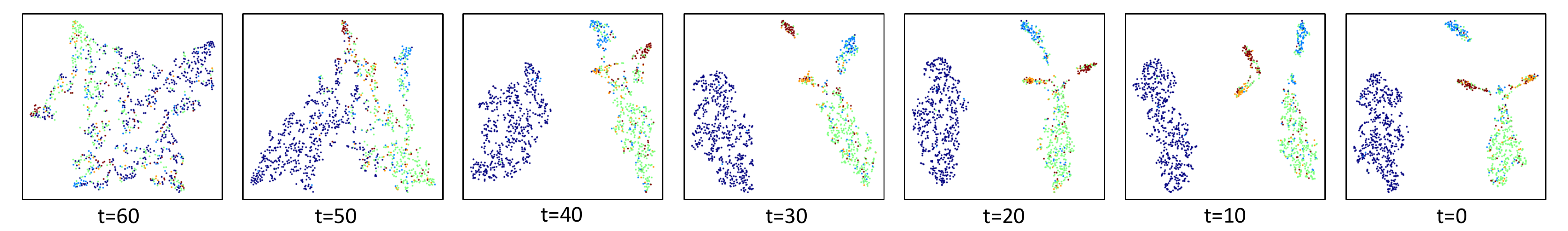}\label{fig: aptos}}}
    \vskip -5pt
    \caption{t-SNE obtained from the denoised feature embedding by the diffusion reverse process during inference on three datasets. $t$ is the current diffusion time step for inference. As the time step encoding progresses, the noise is gradually removed, thereby obtaining a clear distribution of classes; see the last column (please zoom in).}
    \label{fig: vis}
\end{figure*}

\noindent \textbf{Ablation Study:}
Extensive experiments are conducted to evaluate the effectiveness of major modules of our network. To do so, we build three baseline networks from our method.  
The first baseline (denoted as ``basic'') is to remove all diffusion operations and the MMD regularization loss from our network. It means that ``basic'' is equal to the classical ResNet18.
Then, we apply the vanilla diffusion process onto ``basic'' to construct another baseline network (denoted as ``C1''), and further add our dual-granularity conditional guidance into the diffusion process to build a baseline network, which is denoted as ``C2''. Hence, ``C2'' is equal to removing the MMD regularization loss from our network for image classification. 
Table~\ref{tab:ablation} reports the Accuracy and F1-score results of our method and three baseline networks on our PMG2000 dataset. Apparently, compared to ``basic'', ``C1'' has an Accuracy improvement of 0.027 and an F1-score improvement of 0.018, which indicates that the diffusion mechanism can learn more discriminate features for medical image classification, thereby improving the PMG performance. 
Moreover, the better Accuracy and F1-score results of ``C2'' over ``C1'' demonstrates that introducing our dual-granularity conditional guidance into the vanilla diffusion process can benefit the PMG performance.
Furthermore, our method outperforms ``C2'' in terms of Accuracy and F1-score, which indicates that exploring the MMD regularization loss in the diffusion process can further help to enhance the PMG results.

\noindent \textbf{Visualization of our Diffusion Procedure:} 
To illustrate the diffusion reverse process guided by our dual-granularity conditional encoding, we used the t-SNE tool to visualize the denoised feature embeddings at consecutive time steps. Figure~\ref{fig: vis} presents the results of this process on all three datasets. As the time step encoding progresses, the denoise diffusion model gradually removes noise from the feature representation, resulting in a clearer distribution of classes from the Gaussian distribution. The total number of time steps required for inference depends on the complexity of the dataset.

\section{Conclusion}
\label{sec:con}

This work presents a diffusion-based network (DiffMIC) to boost medical image classification. 
The main idea of our DiffMIC is to introduce dual-granularity conditional guidance over vanilla DDPM, and enforce condition-specific MMD regularization to improve classification performance. 
Experimental results on three medical image classification datasets with diverse image modalities show the superior performance of our network over state-of-the-art methods. As the first diffusion-based model for general medical image classification, our DiffMIC has the potential to serve as an essential baseline for future research in this area.

\small {\noindent\textbf{Acknowledgments:} This research is supported by Guangzhou Municipal Science and Technology Project (Grant No. 2023A03J0671), the National Research Foundation, Singapore under its AI Singapore Programme (AISG Award No: AISG2-TC-2021-003), A*STAR AME Programmatic Funding Scheme Under Project A20H4b0141, and A*STAR Central Research Fund. }

%
%
%
\bibliographystyle{splncs04}
\bibliography{paper}

\end{document}